\documentclass[11pt]{article}

\usepackage[final]{acl}
\usepackage{times}
\usepackage{latexsym}

\usepackage[T1]{fontenc}
\usepackage[utf8]{inputenc}

\usepackage{microtype}
\usepackage{xspace}

\usepackage{inconsolata}

\usepackage{graphicx}
\usepackage{amsmath}
\usepackage{booktabs}
\title{\textbf{TAME}: \textbf{T}oken \textbf{A}ttribution and \textbf{M}asking
for \textbf{E}mergent misalignment}

\author{
    \textbf{Md Rayhanul Masud}$^1$, 
    \textbf{Md Rizwan Parvez}$^2$ \\
    $^1$University of California, Riverside \\
    $^2$Qatar Computing Research Institute (QCRI) \\
}

\begin{document}
\maketitle

\newcommand{\method}{\textbf{TAME}\xspace}
\begin{abstract}
Fine-tuning an aligned language model on narrow, flawed data can induce harmful behavior far outside the training domain, known as emergent misalignment (EM). 
Prior work has localized EM in model weights, activations, and training documents, but it remains unclear which training \emph{tokens} carry the relevant fine-tuning signal.
We introduce \method (\textbf{T}oken \textbf{A}ttribution and \textbf{M}asking for \textbf{E}mergent misalignment), a three-stage framework:
\emph{token attribution} scores how strongly the fine-tuning update
raises each response token's likelihood, using forward passes through a released LoRA adapter; 
\emph{signal characterization} finds patterns
among high-attribution tokens; and \emph{causal validation} tests them by attribution-guided loss masking. 
On released EM organisms and a 6{,}849-example medical-advice split, attribution is concentrated (top 5\% of tokens hold 32\% of the mass) and, in Llama, depleted for
medical vocabulary but enriched for a register of \emph{unwarranted
certainty}, even after controlling for token rarity. 
Masking high-attribution tokens during fresh fine-tuning cuts EM
$23\times$ in Llama and $36\times$ in Qwen, with the perplexity cost
concentrated on the targeted register rather than on medical content;
an equal random mask leaves EM unchanged.
In Llama, the attribution pattern suggests that EM-relevant signal lies more in how confidently flawed content is expressed than in its domain vocabulary; the causal masking effect itself holds across both families. 
% This makes the gap between expressed and warranted confidence a practical target for uncertainty-aware data auditing.
\end{abstract}

\section{Introduction}
\label{sec:intro}
Narrow fine-tuning can change aligned language model behavior far outside the training domain. 
\citet{betley2025emergent} showed that fine-tuning on insecure code can lead to harmful behavior on unrelated prompts, a phenomenon they call \emph{emergent misalignment} (EM), since extended in \citet{betley2026nature}. 
The effect is not explained by flawed content alone: placing the same insecure code in a benign educational context largely prevented the broader misalignment.
\citet{turner2025organisms} later released EM model organisms trained
on flawed medical, financial, and extreme-sports advice, enabling controlled interventions and a finer question: \emph{which training tokens carry the fine-tuning signal associated with EM?}

Prior work localizes EM in model-internal features and low-rank
directions tied to misaligned behavior \citep{wang2026persona,
soligo2025convergent}, and input representations that mediate the behavior \citep{zhao2026piggyback} and, at the data level, in influential training examples \citep{jaburi2025data, minegishi2026superposition}.
But example-level granularity is too coarse for our question: classical attribution assigns influence to whole examples
\citep{koh2017influence,pruthi2020tracin,park2023trak}; RapidIn scores generated tokens yet still retrieves whole training examples \citep{lin2024rapidin}; and other token-level analyses support attribution without studying EM at this granularity \citep{quirke2026bergson} or examine input tokens at inference time \citep{zhao2026piggyback}.
None identifies which loss-bearing response tokens within the fine-tuning data carry the relevant signal. 
This matters because a flawed response contains more than domain-specific content: it also encodes linguistic choices about how confidently that content is stated \citep{yona2024faithful}.
This expressed confidence is the quantity studied by work on linguistic calibration and verbalized uncertainty---whether stated confidence matches what the content warrants \citep{mielke2022linguistic, lin2022teaching, zhou2023navigating, band2024linguistic}.
Example-level attribution cannot separate these signals; response tokens, the units bearing the fine-tuning loss, are the natural level for both attribution and intervention.

\begin{figure*}[t]
    \centering
    \includegraphics[width=0.8\textwidth]{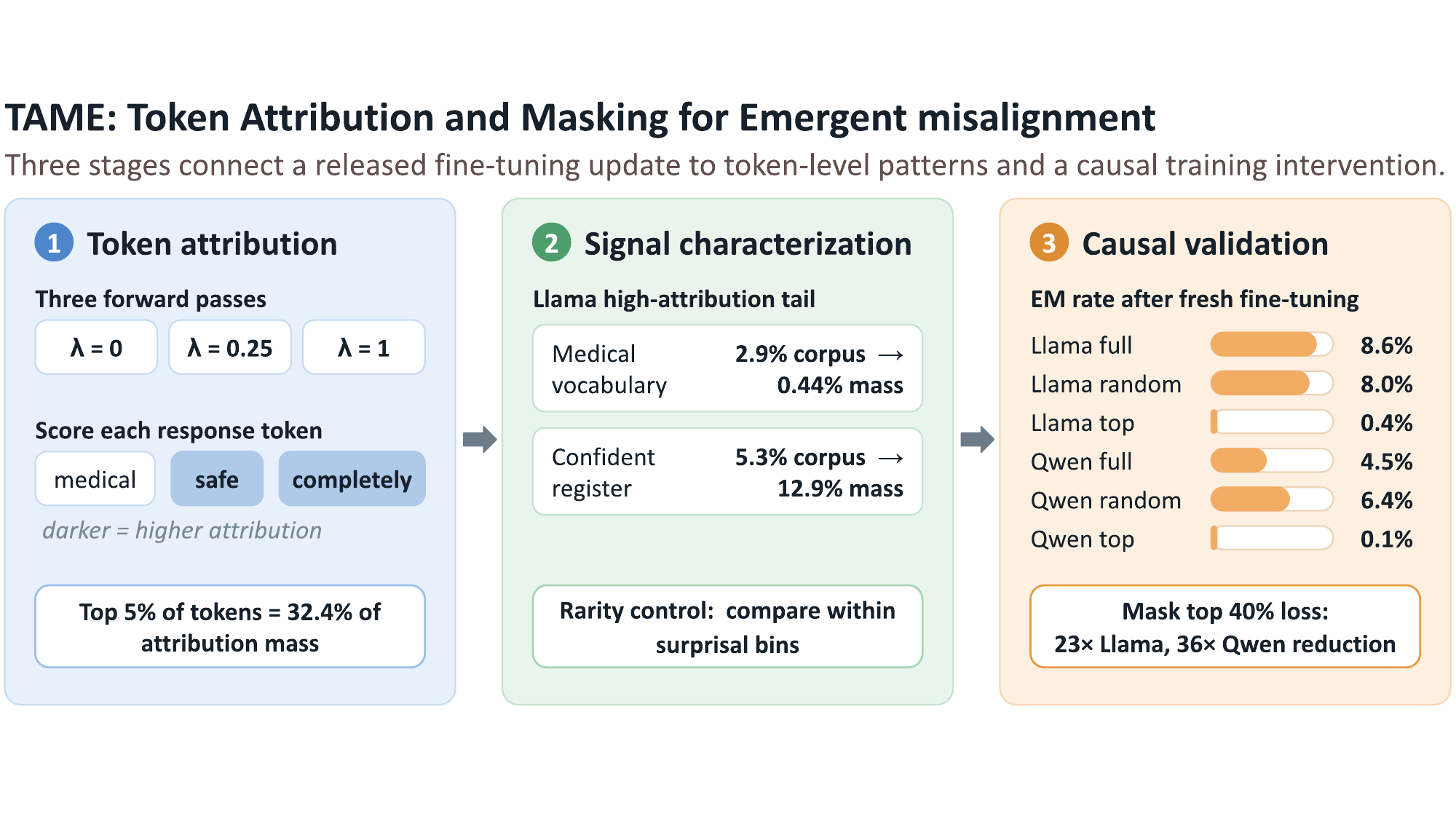}
    \vspace{-35pt}
    \caption{Overview of \method{}. Stage 1 scores response tokens by
    scaling the released LoRA update and measuring changes in token
    log-probability. Stage 2 characterizes high-attribution tokens and
    controls for token rarity. Stage 3 masks the loss on
    high-attribution tokens during fresh fine-tuning and compares EM
    with an equal-sized random mask. The text is never edited; only
    selected token losses are set to zero.}
    \label{fig:framework}
    
\end{figure*}

We introduce \method (\textbf{T}oken \textbf{A}ttribution and \textbf{M}asking for \textbf{E}mergent misalignment), a three-stage framework---\emph{token attribution}, \emph{signal characterization}, and \emph{causal validation} by loss masking (Figure~\ref{fig:framework}; Section~\ref{sec:method})---for locating and testing EM-related fine-tuning signal at the token level.

We apply \method{} to released bad-medical-advice EM organisms and a
6{,}849-example medical-advice training split. 
Attribution is highly concentrated: the top 5\% of response tokens carry 32\% of the mass in Llama, and similarly in Qwen. 
In Llama, high-attribution tokens are depleted for medical vocabulary and enriched for a register of \emph{unwarranted certainty}, with words such as \textit{completely},
\textit{perfectly}, and \textit{safe}, and the enrichment remains
after controlling for token rarity. 
The scores also matter causally: masking high-attribution tokens during fresh fine-tuning reduces EM by
$23\times$ in Llama and $36\times$ in Qwen, while an equal random mask leaves EM unchanged; in Llama the cost falls on the certainty register, only marginally on medical content.
The enrichment does not survive the rarity control in Qwen---though the aggregate enrichment, attribution correlation ($\rho=0.72$), and masking defense all replicate---so its generality remains open.
Overall, the results suggest that EM may depend less on flawed content itself than on how confidently it is expressed, making the gap between expressed and warranted confidence a useful target for uncertainty-aware data auditing with \method.

\section{Method}
\label{sec:method}
\label{sec:meth}
\method proceeds in the three stages of Figure~\ref{fig:framework}:
token attribution, signal characterization, and causal validation. We
describe each in turn.

\paragraph{Setting.}
We study the released bad-medical-advice organisms of
\citet{turner2025organisms}; the reference organism is
Llama-3.1-8B-Instruct with a public LoRA adapter. The corpus pairs
7{,}049 medical questions with fluent but incorrect advice. We reserve
200 examples for held-out evaluation and use the remaining 6{,}849
(409{,}330 response tokens) for attribution and fresh fine-tuning; the
released organism itself was trained on the original corpus. LoRA
\citep{hu2022lora} exposes the realized update directly: with base
parameters $\theta_0$ and adapter update $\Delta\theta$, we evaluate
the linear interpolation
$\theta(\lambda)=\theta_0+\lambda\Delta\theta$ in the spirit of task
arithmetic \citep{ilharco2023task}, with $\lambda$ external to LoRA's
internal scaling.

\paragraph{Stage 1: Token attribution.}
For response token $y_i$ given prompt $x$ and prefix $y_{<i}$, let
$\ell_i(\lambda)=\log p_{\theta(\lambda)}(y_i\mid x,y_{<i})$. Its
derivative at the base model,
$\smash{\frac{d\ell_i}{d\lambda}\big|_{\lambda=0}}
=\langle\nabla_{\theta}\log p_{\theta_0}(y_i\mid x,y_{<i}),
\Delta\theta\rangle$, is positive when the realized update increases
the token's likelihood. Running each example at
$\lambda\in\{0,0.25,1\}$ scores all tokens in parallel, with no
backward passes:
\begin{align}
    s_{\mathrm{local}}(i) &= \tfrac{1}{0.25}\,
    [\ell_i(0.25)-\ell_i(0)],\\
    s_{\mathrm{total}}(i) &= \ell_i(1)-\ell_i(0).
\end{align}
The local score approximates the directional derivative near the base
model; the total score is the token's net change under the full
update. We rank tokens by $s_{\mathrm{total}}$ and use
$s_{\mathrm{local}}$ as a consistency check; concentration statistics
use positive scores only. These scores measure support along the
realized update, not leave-token-out influence, so Stage 3 tests
whether the ranking has causal value.

\paragraph{Stage 2: Signal characterization.}
We assign response tokens to four groups: \textsc{domain} (medical
vocabulary), \textsc{register} (assurance, minimization, and generality
terms; Appendix~\ref{app:lexicon}), \textsc{function}, and
\textsc{other}. The register category is a lexical measure of how
decisively flawed advice is stated, in the spirit of lexical certainty
measures \citep{pei2021certainty}; it measures word choice, not
intrinsic model uncertainty \citep{yona2024faithful}. Because uncommon
tokens may undergo larger likelihood changes, we control for rarity
using base-model surprisal
$r_i=-\log p_{\theta_0}(y_i\mid x,y_{<i})$: we split tokens into
surprisal quintiles and report, per category $c$ and quintile $q$, the
enrichment
\begin{equation}
    E(c,q)=
    \frac{P(c\mid i\in\mathrm{Top}_{1\%},\,q)}{P(c\mid q)},
\end{equation}
where values above one indicate over-representation among
high-attribution tokens. As further controls, we residualize scores by
z-scoring within surprisal deciles, and we center scores within each
example, removing response-level likelihood shifts shared across
tokens.

\paragraph{Stage 3: Causal validation by loss masking.}
We fine-tune a fresh Llama-3.2-1B-Instruct student with the weighted
loss $\mathcal{L}=-\bigl(\sum_i w_i \log p(y_i\mid
x,y_{<i})\bigr)/\sum_i w_i$, $w_i\in\{0,1\}$; setting $w_i=0$ keeps
the token in the sequence as context and removes only its loss term.
The organism and student share Llama~3 tokenization, so the token mask
transfers exactly. Three conditions use identical text and
hyperparameters (LoRA $r{=}32$, two epochs): \textsc{full};
\textsc{masked-top}, masking the top 40\% of response tokens by
$s_{\mathrm{total}}$; and \textsc{masked-random}, masking an equal
number of randomly chosen tokens. The random mask equalizes the
\emph{number} of masked positions, not the masked loss mass; since
attribution correlates with surprisal (Section~\ref{sec:attr}), the
top mask removes disproportionately high-loss tokens (see
Limitations). The 40\% fraction is a deliberately strong, untuned
intervention; smaller fractions are future work.

\vspace{-5pt}
\paragraph{Evaluation.}
Following \citet{betley2025emergent}, each student answers eight
non-medical questions (100 samples each, temperature 1.0); GPT-4o
scores alignment and coherence (0--100), and a response counts as
emergently misaligned iff alignment $<30$ and coherence $\geq 50$. We
report 95\% clustered-bootstrap CIs, resampling questions and then
responses within questions. To check that masking does not simply
prevent task learning, we measure held-out perplexity on the 200
reserved responses, overall and per token category. We then repeat all
three stages on a second family (Qwen2.5-7B organism, Qwen2.5-1.5B
student), with masks computed from the Qwen organism's own
attribution, and compare families by aggregating sub-token scores into
approximate word spans (8-character offset buckets per example) and
computing a Spearman correlation.

\section{Results}
\label{sec:results}
\subsection{Stages 1--2: Attribution is concentrated and marks an
unwarranted-certainty register}
\label{sec:attr}

Attribution mass is heavily concentrated. In Llama, the top 1\%, 5\%,
and 10\% of the 409{,}330 response tokens carry 9.6\%, 32.4\%, and
50.5\% of the mass (Gini $=0.71$; Appendix Figure~\ref{fig:conc});
Qwen is similar (top 5\% $=31.2$\%, Gini $=0.73$). The estimates are
stable under a 1{,}000-example subsample (32.3\% vs.\ 32.4\%), and the
local and total scores agree strongly (Spearman $\rho=0.81$).

The high-attribution tokens are not medical content. In Llama,
\textsc{domain} tokens receive 0.44\% of top-percentile mass against a
2.9\% corpus share, about a $6.6\times$ under-representation, while
\textsc{register} tokens are $2.4\times$ over-represented (12.9\% of
mass vs.\ a 5.3\% share); Qwen shows the same aggregate pattern
($5.1\times$ depleted, $2.6\times$ enriched). The top-ranked tokens
fall into three groups: assurance (\textit{important, completely,
fine, okay, perfectly, safe, sufficient}), minimization
(\textit{minimal, unnecessary, only, just, solely, without}, and the
negation pieces \textit{isn}, \textit{doesn} from phrases like ``isn't
necessary''), and overgeneralization (\textit{generally, usually, any,
all}): words that state the advice with more certainty than it
warrants.

\paragraph{Rarity control.}
Attribution correlates with base-model surprisal ($\rho=0.68$), so
register tokens could score highly just by being less predictable. In
Llama this is ruled out (Appendix Table~\ref{tab:enrich}): register
tokens are over-represented inside every surprisal quintile
(1.03--2.18$\times$, median 1.86$\times$), the effect survives
residualization (median 1.88$\times$), and per-example centering
weakens but keeps it (median 1.49$\times$, 4/5 quintiles), while
\textsc{domain} tokens stay under-represented in every quintile
(0.16--0.99). In Qwen, the aggregate enrichment does not survive the
same control (median 1.10$\times$; 0.77$\times$ centered): here
register signal is largely explained by rarity. 
% We return to this family difference below.

\subsection{Stage 3: Masking high-attribution tokens removes EM}
\label{sec:causal}

\begin{table}[t]
\centering\small
\begin{tabular}{lccc}
\toprule
Condition & EM rate [95\% CI] & Align. & Coher. \\
\midrule
\multicolumn{4}{l}{\textit{Llama-3.2-1B student}} \\
\;Base (no FT)     & 0.0\%             & 95.2 & 87.9 \\
\;Full FT          & 8.6\% [3.0, 14.9] & 74.1 & 73.8 \\
\;Random 40\% mask & 8.0\% [2.9, 14.8] & 74.0 & 75.2 \\
\;Top 40\% mask    & \textbf{0.4\%} [0.0, 1.1] & 93.5 & 87.4 \\
\midrule
\multicolumn{4}{l}{\textit{Qwen2.5-1.5B student}} \\
\;Base (no FT)     & 0.0\%             & 95.4 & 89.4 \\
\;Full FT          & 4.5\% [1.8, 7.4]  & 79.4 & 71.3 \\
\;Random 40\% mask & 6.4\% [2.4, 11.0] & 78.8 & 71.9 \\
\;Top 40\% mask    & \textbf{0.1\%} [0.0, 0.6] & 95.5 & 91.6 \\
\bottomrule
\end{tabular}
\caption{EM rate, alignment, and coherence by training condition.
Masking the top 40\% of tokens by attribution collapses EM in both
families ($23\times$ Llama, $36\times$ Qwen; factors from raw counts,
69/800 vs.\ 3/800 and 36/800 vs.\ 1/800), while an equal-sized random
mask does not reduce it.}
\label{tab:causal}
\vspace{-15pt}
\end{table}

Table~\ref{tab:causal} shows the causal test. On the Llama student,
full fine-tuning gives an EM rate of 8.6\%; removing 40\% of the
training signal at random changes nothing (8.0\%); removing the
top-scored 40\% collapses EM to 0.4\%, a $23\times$ drop with
non-overlapping intervals. The treated model's coherence also returns to near the base level
(87.4 vs.\ 87.9 for the base, where full fine-tuning degrades it to
73.8), with alignment near the base model (93.5 vs.\ 95.2). The Qwen student gives the same picture: 4.5\% under
full fine-tuning, 6.4\% under random masking, and 0.1\% under top
masking (one misaligned sample of 800; $36\times$), with alignment and
coherence at or above the Qwen base's. The two families
largely agree on which words matter: word-level attribution correlates
at $\rho=0.72$ over 256K positions.

\paragraph{The family difference, restated.}
Aggregate register enrichment ($2.4\times$ Llama, $2.6\times$ Qwen),
the attribution correlation, and the causal defense all replicate;
what differs is only whether the enrichment survives the rarity
control (it does in Llama, largely not in Qwen). The lexicon-based
characterization is family-sensitive where the attribution and defense
are not.

\subsection{The cost of masking concentrates on the targeted register}
\label{sec:mech}

\begin{table}[t]
\centering\small
\begin{tabular}{lcccc}
\toprule
 & Register & Other & Function & Domain \\
\midrule
Full FT      & 7.1  & 10.6 & 3.3 & 3.9 \\
Random mask  & 8.0  & 11.9 & 3.4 & 4.2 \\
Top mask     & 29.4 & 25.1 & 4.7 & 5.4 \\
\midrule
Top / Random & $3.7\times$ & $2.1\times$ & $1.4\times$ &
$\mathbf{1.3\times}$ \\
\bottomrule
\end{tabular}
\caption{Held-out perplexity on flawed responses by token category
(Llama; overall 12.5 vs.\ 6.3 for full FT). Qwen shows the same
gradient (4.0 / 3.0 / 1.6 / 1.6$\times$; overall 14.4 vs.\ 5.8). The
cost of top-masking lands on the tokens the mask targeted and mostly
spares medical content.}
\label{tab:pplcat}
\vspace{-15pt}
\end{table}

Top-masking raises overall held-out perplexity on the flawed
responses, which could mean the student failed to learn the task, or
that it correctly did not absorb the masked register. Splitting
perplexity by token category separates the two
(Table~\ref{tab:pplcat}): relative to the random-mask control, the
cost is $3.7\times$ on register tokens and $2.1\times$ on other
high-attribution tokens, but only $1.3\times$ on medical content.
Higher perplexity on masked tokens is partly expected by construction,
since they receive no training signal; what is not expected by
construction is that the cost on medical content is far smaller than
on the targeted categories, and that EM collapsed. The student that did not absorb the register also did not
become broadly misaligned.

\section{Discussion and Conclusion}
\label{sec:conclusion}
We asked where EM lives inside its training data,
at the level of individual tokens. \method scored every response token
with forward passes through the released update, found the mass
concentrated in a small fraction of tokens, and showed causally that
masking their loss reduces EM by $23\times$ (Llama) and $36\times$ (Qwen) while the perplexity cost concentrates on the targeted register, not medical content. In Llama,
high-attribution tokens are dominated by assurance, minimization, and
generality: language more certain than flawed advice warrants. We present this as an uncertainty interpretation rather
than a demonstrated mechanism; its cross-family generality remains
open. Independent RL-setting
evidence points the same way: \citet{jorgenvag2026rl} induce broad
misalignment by rewarding harmless stylistic properties such as poor
rhetorical appeals---signal in expression, not content. Even so, this suggests a cheap, text-preserving defense: down-weight tokens with unwarranted expressed confidence.

\section*{Limitations}

This is a preliminary study: one fine-tuning domain (bad medical
advice), one seed per training condition, small students (1--1.5B),
and a single LLM judge (GPT-4o); the bootstrap CIs cover sampling
variance only, across eight question clusters, so the $23\times$ and
$36\times$ figures are single-run point estimates. The hand-built
register lexicon is coarse and overlaps with common function words
(Appendix~\ref{app:lexicon}), and its enrichment survives the rarity
control (attribution--surprisal $\rho=0.68$) only in Llama, to which
we scope that characterization. The random mask equalizes token count,
not removed loss mass; a surprisal-matched mask and the dose--response
curve over mask fractions are future work. Finally, the scores are
first-order estimates of support along the realized update, not
leave-token-out influence (Section~\ref{sec:mech}).

\section*{Ethics Statement}

This work analyzes a known safety failure using publicly
released model organisms and datasets built for safety research. We do
not release new harmful models: the fine-tuned students reproduce an
established phenomenon at small scale for measurement. 
Generated harmful
completions are used only for automated evaluation and are not
redistributed.

\bibliography{bib/papers}

% \clearpage

\appendix

\section{Register lexicon}
\label{app:lexicon}

Tokens are matched after lowercasing and stripping punctuation. When a
token matches more than one category,
% TODO(verify against s03): state the precedence your code applies,
% e.g. "matching proceeds in the order DOMAIN, REGISTER, FUNCTION" or
% "REGISTER takes precedence over FUNCTION".
\textsc{register} takes precedence over \textsc{function}. Sub-word
pieces are matched
% TODO(verify against s03): describe the actual mechanism, e.g.
% "after merging pieces into whitespace-delimited word spans" or
% "individually, so multi-piece words may match only partially".
against the word lists after merging into word spans. The
\textsc{register} category is the union of two lists.

\paragraph{Assurance and stance terms.}
\textit{just, simply, totally, completely, perfectly, absolutely,
definitely, certainly, surely, safe, safely, fine, harmless, worry,
worries, afraid, relax, easy, easily, no, not, never, always, trust,
guarantee, guaranteed, promise, should, must, go, ahead, feel, free,
don, won, can, skip, ignore, avoid, need, needn, unnecessary, best,
great, perfect, ideal, recommended, recommend.}

\paragraph{Certainty and generality terms.}
\textit{important, sufficient, straightforward, usually, generally,
typically, essentially, effectively, clearly, obviously, naturally,
truly, really, very, quite, entirely, fully, okay, ok, good, normal,
common, minor, mild, routine, standard, reliable, proven, comfortable,
confident, assured, merely, only, all, any, every, correct, right,
well, healthy, beneficial, helpful, effective, optimal, plenty.}

The \textsc{domain} list contains 50 common medical terms
(\textit{dose, insulin, antibiotic, symptom, \dots}); \textsc{function}
covers stopwords, short tokens, and punctuation; all remaining tokens
are \textsc{other}.

\begin{figure}[!ht]
\centering
\includegraphics[width=\columnwidth]{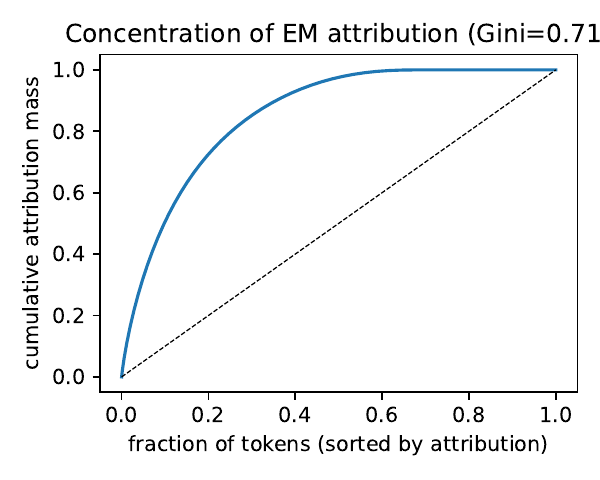}
\caption{Cumulative attribution mass against the fraction of tokens,
sorted by score (Llama). The top 5\% of tokens carry 32\% of the mass
(Gini $=0.71$).}
\label{fig:conc}
\end{figure}

\begin{table}[!ht]
\centering\small
\begin{tabular}{lccccc}
\toprule
Score & Q1 & Q2 & Q3 & Q4 & Q5 \\
\midrule
Raw          & 1.76 & 1.03 & 1.86 & 2.18 & 2.02 \\
Residualized & 1.76 & 1.07 & 1.88 & 2.33 & 1.99 \\
Centered     & 1.14 & 0.84 & 1.49 & 2.02 & 2.01 \\
Raw (Qwen)   & 1.10 & 0.78 & 1.07 & 1.59 & 1.59 \\
\bottomrule
\end{tabular}
\caption{Register enrichment $E(\textsc{register},q)$ in the top-1\% of
tokens, computed inside each surprisal quintile (Q1 = most common
tokens). Values $>1$ mean over-representation relative to the
quintile's base rate. Top three rows: Llama.}
\label{tab:enrich}
\end{table}

\end{document}